\documentclass[conference]{IEEEtran}
\IEEEoverridecommandlockouts
\usepackage{cite}
\usepackage{amsmath,amssymb,amsfonts}
\usepackage{algorithmic}
\usepackage{graphicx}
\usepackage{graphics}
\usepackage{textcomp}
\usepackage{xcolor}
\usepackage{tabularx}   
\usepackage{makecell}   
\usepackage{booktabs}   
\usepackage{multirow}
\usepackage{array}
\usepackage{tablefootnote}
\usepackage[ruled,vlined,linesnumbered]{algorithm2e}
\SetAlFnt{\footnotesize}
\SetAlCapFnt{\footnotesize}
\SetAlCapNameFnt{\footnotesize}
\SetAlCapNameSty{textbf}
\SetAlTitleFnt{\footnotesize}
\SetKwInput{KwInput}{Input}
\SetKwInput{KwOutput}{Output}
\SetKwInput{KwParams}{Parameters}
\SetKwInput{KwReturn}{Return}
\newcolumntype{C}[1]{>{\centering\arraybackslash}p{#1}}\newcolumntype{L}[1]{>{\raggedright\let\newline\\\arraybackslash\hspace{0pt}}m{#1}}
\newcolumntype{R}[1]{>{\raggedleft\let\newline\\\arraybackslash\hspace{0pt}}m{#1}}
\newcolumntype{Y}{>{\centering\arraybackslash}X}
\usepackage[super]{nth}
\def\BibTeX{{\rm B\kern-.05em{\sc i\kern-.025em b}\kern-.08em
    T\kern-.1667em\lower.7ex\hbox{E}\kern-.125emX}}
\begin{document}
\makeatletter 
\newcommand{\linebreakand}{%
  \end{@IEEEauthorhalign}
  \hfill\mbox{}\par
  \mbox{}\hfill\begin{@IEEEauthorhalign}
}
\makeatother 

\title{Meta--cognitive Multi--scale Hierarchical Reasoning for Motor Imagery Decoding
\footnote{{\thanks{This research was supported by the Institute of Information \& Communications Technology Planning \& Evaluation (IITP) grant, funded by the Korea government (MSIT) (No. RS-2019-II190079, Artificial Intelligence Graduate School Program (Korea University) and  No. IITP-2025-RS-2024-00436857, Information Technology Research Center (ITRC)).}
}}
}

\author{
\IEEEauthorblockN{Si--Hyun Kim}
\IEEEauthorblockA{\textit{Dept. of Artificial Intelligence} \\
\textit{Korea University} \\
Seoul, Republic of Korea \\
kim\_sh@korea.ac.kr} \\
Byoung--Hee Kwon \\
\textit{Dept. of Brain and Cognitive Engineering} \\
\textit{Korea University} \\
Seoul, Republic of Korea \\
bh\_kwon@korea.ac.kr

\and 

\IEEEauthorblockN{Heon--Gyu Kwak}
\IEEEauthorblockA{\textit{Dept. of Artificial Intelligence} \\
\textit{Korea University} \\
Seoul, Republic of Korea \\
hg\_kwak@korea.ac.kr} \\
Seong--Whan Lee \\
\textit{Dept. of Artificial Intelligence} \\
\textit{Korea University} \\
Seoul, Republic of Korea \\
sw.lee@korea.ac.kr
}

\maketitle 

\begin{abstract}
Brain--computer interface (BCI) aims to decode motor intent from noninvasive neural signals to enable control of external devices, but practical deployment remains limited by noise and variability in motor imagery (MI)--based electroencephalogram (EEG) signals. This work investigates a hierarchical and meta--cognitive decoding framework for four--class MI classification. We introduce a multi--scale hierarchical signal processing module that reorganizes backbone features into temporal multi--scale representations, together with an introspective uncertainty estimation module that assigns per--cycle reliability scores and guides iterative refinement. We instantiate this framework on three standard EEG backbones (EEGNet, ShallowConvNet, and DeepConvNet) and evaluate four--class MI decoding using the BCI Competition IV--2a dataset under a subject--independent setting. Across all backbones, the proposed components improve average classification accuracy and reduce inter--subject variance compared to the corresponding baselines, indicating increased robustness to subject heterogeneity and noisy trials. These results suggest that combining hierarchical multi--scale processing with introspective confidence estimation can enhance the reliability of MI--based BCI systems.
\end{abstract}


\section{INTRODUCTION}
Brain--computer interface (BCI) systems establish a direct communication pathway between the human brain and external devices and are actively explored for assistive technology and human--computer interaction \cite{yin2014dynamically}. Among non-invasive paradigms, motor imagery (MI) has been extensively investigated owing to its high potential for practical applications. MI is the mental simulation of movement without overt execution and elicits discriminative activity in sensorimotor cortex. Decoding these patterns from electroencephalogram (EEG) signals enables intuitive control of effectors such as wheelchairs \cite{cho2021neurograsp} and robotic arms \cite{zhou2023shared}.

Despite this promise, deploying MI--based BCI systems in practice remains difficult. MI--based EEG decoding is hindered by low signal--to--noise ratio and physiological artifacts \cite{prabhakar2020framework}, pronounced inter-- and intra--subject variability \cite{suk2014predicting}, non--stationarity over time \cite{huang2021review}, and complex changes in functional connectivity \cite{ding2013changes}. The conventional methods, such as subject-- and class--specific frequency band selection \cite{suk2011subject}, classical pattern recognition and machine learning models \cite{lotte2018review,lee2003pattern}, and deep learning--based decoders including compact convolutional architectures \cite{lawhern2018eegnet,schirrmeister2017deep} and models for continuous mental state decoding \cite{lee2020continuous}, still struggle to deliver the reliability required for real--world use, often degrading on new or noisy trials. This calls for models that capture multi--scale dynamics while providing reliable predictions.

\begin{figure*}[ht]
    \centering
\includegraphics[width=16.5cm]{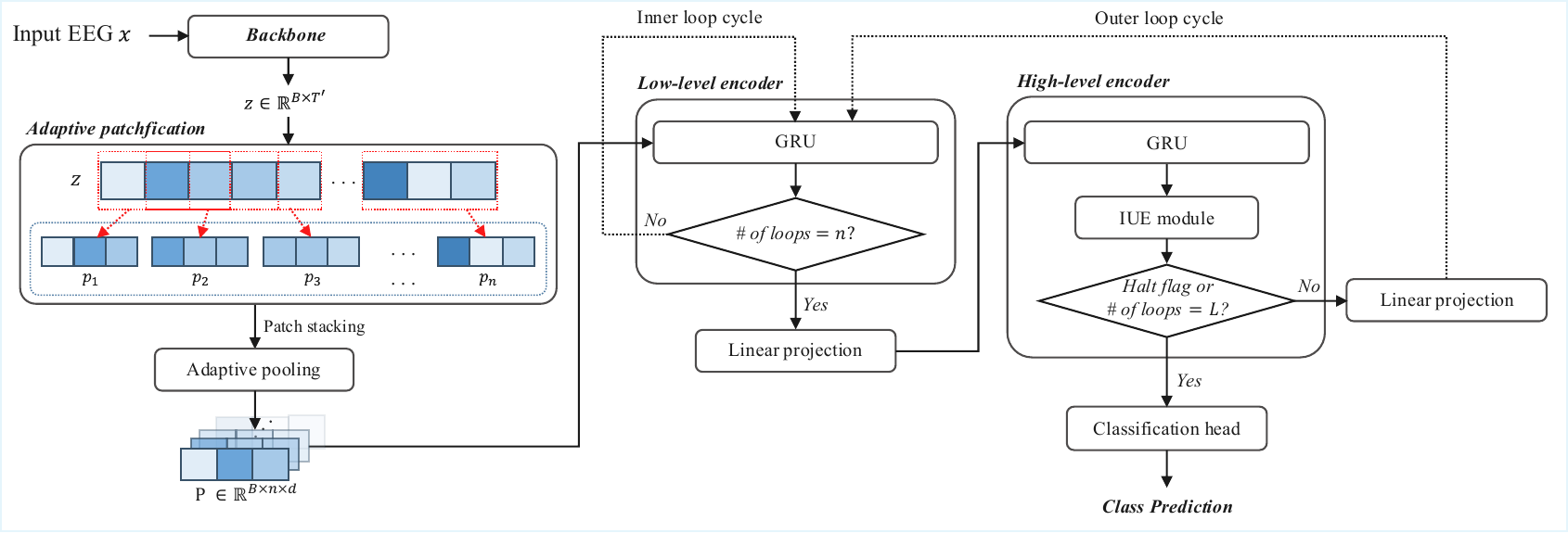}
    \caption{
    Overview of the proposed framework. Backbone features are partitioned into adaptive windows and patches. A low--level GRU with top--down gating summarizes patches per window, a high--level GRU integrates across windows to yield per--cycle logits and a halting score.
    }
    \label{fig:fig1}
\end{figure*}

To address these challenges, we leverage inspiration from hierarchical processing and meta-cognitive reasoning principles that have proven effective in related domains. On studies related to the hierarchical processing, classical models and early neural architectures for visual pattern recognition, including biologically inspired convolutional models such as the neocognitron, decompose inputs into coordinated sub-tasks or multi-resolution representations \cite{lee1997new,fukushima1980neocognitron}. Multilayer cluster neural networks demonstrate how clustered representations support robust pattern recognition \cite{lee1995multilayer}, while integrated segmentation and recognition of handwritten numerals highlight cascaded processing pipelines \cite{lee1999integrated}. Classical multi-resolution analysis with wavelet transforms provides a framework for hierarchical signal decomposition across scales \cite{mallat2002theory}, and multi-resolution recognition models emphasize processing structure at multiple spatial scales \cite{lee1996multiresolution}. Also spatio--temporal activity analysis extends these ideas to motion patterns over time \cite{lee2015motion}. Modern deep learning methods, such as diffusion--based hierarchical planners and deep reinforcement learning, explicitly separate high--level intent from low--level execution across temporal horizons \cite{chen2024simple,lee2018deep}.

Studies inspired by meta--cognitive reasoning, on the other hand, demonstrated the effectiveness of controlling the learning and inference process via self--evaluation. Agent--level frameworks such as Reflexion steer multi--step reasoning using internal confidence and reflective feedback \cite{shinn2023reflexion}, and recent large language model frameworks autonomously construct and adapt multi--stage reasoning procedures such as chain--of--though, highlighting meta--cognitive mechanisms that decide when to trust, refine, or revise intermediate predictions. These hierarchical and introspective approaches resonate with neural mechanism integrate information across scales into abstract, modality--invariant internal representations \cite{kim2015abstract,bulthoff2003biologically}.

Motivated by these trends, we propose a multi--scale hierarchical signal processing (MHSP) module and an introspective uncertainty estimation (IUE) head for decoding MI--based EEG signals. The MHSP module reorganizes backbone features into multi--scale temporal windows and summarizes them through a two--level recurrent hierarchy, while the IUE head provides meta--cognitive reliability estimates over iterative refinement cycles. The proposed method showed that combining hierarchical abstraction with meta--cognitive uncertainty estimation yields more robust decoding under noisy EEG conditions and improves reliability of MI--based BCI systems. \\

\section{METHODS}
We extend three EEG decoding backbones with an MHSP module and an IUE module for subject--independent MI classification, drawing on hierarchical and meta--cognitive design principles \cite{wang2025hierarchical, zhang2025stair}.
\subsection{MHSP Module}
Given an input EEG trial $x\in \mathbb{R}^{C \times T}$, a backbone produces a feature map $z\in \mathbb{R}^{B \times T'}$, where $B$ is the batch size, $T'$ is the output feature size. MHSP then segments $z$ into patches, using a sliding window of size $w$. The resulting stacks of patches are then passed through low-- and high--level encoders, which recurrently prune less informative patches across recursive refinement cycles.

\begin{algorithm}[]
\setlength{\lineskip}{3.5pt}
\caption{Low--level patch refinement encoder}
\DontPrintSemicolon
\KwInput{Patch series $P=\{p_t\}_{t=1}^{n}$}
\KwParams{$W_z,W_r,W_h\in\mathbb{R}^{d_h\times d_x}$, 
$U_z,U_r,U_h\in\mathbb{R}^{d_h\times d_h}$, 
$b_z,b_r,b_h\in\mathbb{R}^{d_h}$, 
$\gamma,\beta\in\mathbb{R}^{d_h}$}
\KwOutput{Refined feature patches $h_{LLE}\in\mathbb{R}^{d_h}$}

$h_0 \leftarrow \mathbf{0}$\  \tcp*{initial state}
\For{$t \leftarrow 1$ \KwTo $n$}{
  $z_t \leftarrow \sigma(W_z p_t + U_z h_{t-1} + b_z)$ \tcp*{update gate}
  $r_t \leftarrow \sigma(W_r p_t + U_r h_{t-1} + b_r)$ \tcp*{reset gate}
  $\tilde{h}_t \leftarrow \tanh(W_h p_t + U_h (r_t \odot h_{t-1}) + b_h)$ \tcp*{candidate}
  $h_t \leftarrow (1 - z_t)\odot h_{t-1} + z_t\odot \tilde{h}_t$ \tcp*{next hidden state}
}
$r \leftarrow \sqrt{\frac{1}{d_h}\sum_{i=1}^{d_h}(h_n)_i^2 + \varepsilon}$ \tcp*{normalized $h_n$}
$h_{LLE} \leftarrow \gamma \odot \frac{h_n}{r} + \beta$

\KwReturn{$h_{LLE}$}
\end{algorithm}

\subsubsection{Adaptive patchfication}
We patch $z$ along with the $T'$ dimension with overlapping windows of size $w$, resulting in patches $\{p_1, p_2, \dots ,p_n\}$, which represent the local dynamics of $z$ for each. The patches are then stacked into the patch series $P\in\mathbb{R}^{B \times n \times w}$ and passed through an adaptive pooling layer to align the patch dimensions to $d$ before feeding them into hierarchical encoders to capture the contextual evolution across patches.

\begin{table*}[t!]
\centering
\caption{Overall performance across subjects for baseline and proposed models incorporating MHSP module and IUE module.}
\renewcommand{\arraystretch}{1}
\setlength{\tabcolsep}{4pt}

\begin{tabular}{c|ccc|ccc|ccc}
\toprule
\multicolumn{1}{c|}{\multirow[c]{2}{*}{\begin{tabular}{c}Subject\end{tabular}}}
& \raisebox{-1ex}{EEGNet}
& \makecell[t]{EEGNet\\w/ MHSP}
& \makecell[t]{EEGNet\\w/ MHSP+IUE}
& \makecell[t]{Shallow\\ConvNet}
& \makecell[t]{ShallowConvNet\\w/ MHSP}
& \makecell[t]{ShallowConvNet\\w/ MHSP+IUE}
& \makecell[t]{Deep\\ConvNet}
& \makecell[t]{DeepConvNet\\w/ MHSP}
& \makecell[t]{DeepConvNet\\w/ MHSP+IUE} \\
\midrule
S1 & 0.532 & 0.519 & 0.533  & 0.625 & 0.486 & 0.521 & 0.560 & 0.537 & 0.524 \\
S2 & 0.407 & 0.532 & 0.554 & 0.361 & 0.366 & 0.388 & 0.577 & 0.472 & 0.520 \\
S3 & 0.583 & 0.699 & 0.685 & 0.616 & 0.574 & 0.621 & 0.634 & 0.625 & 0.642 \\
S4 & 0.440 & 0.463 & 0.472 & 0.394 & 0.398 & 0.402 & 0.431 & 0.444 & 0.465 \\
S5 & 0.435 & 0.588 & 0.575 & 0.343 & 0.458 & 0.463 & 0.472 & 0.551 & 0.566 \\
S6 & 0.463 & 0.537 & 0.545 & 0.384 & 0.403 & 0.417 & 0.491 & 0.523 & 0.532 \\
S7 & 0.560 & 0.681 & 0.686 & 0.482 & 0.560 & 0.594 & 0.607 & 0.583 & 0.601 \\
S8 & 0.741 & 0.676 & 0.680 & 0.625 & 0.653 & 0.642 & 0.602 & 0.662 & 0.671 \\
S9 & 0.583 & 0.607 & 0.621 & 0.657 & 0.671 & 0.664 & 0.690 & 0.676 & 0.691 \\
\midrule
Acc. & 0.527 & 0.589 & 0.592 & 0.499 & 0.508 & 0.512 & 0.551 & 0.564 & 0.568 \\
Std.    & 0.099 & 0.078  & 0.070 & 0.124 & 0.106 & 0.112 & 0.083 & 0.080 & 0.077 \\
\bottomrule
\end{tabular}

\vspace{2mm}
\begin{minipage}{\textwidth}
\raggedright
\quad \quad \footnotesize $^*$Acc.: average accuracy, $^*$Std.: standard deviation, $^*$w/: with
\end{minipage}
\end{table*}

\subsubsection{Multi--scale hierarchical encoders}
The patch series $P$ passes through low--level encoder (LLE) first, which is implemented with gated-recurrent units (GRU)\cite{chung2014empirical} and operated with the Algorithm 1.
The output $h_{LLE}$ corresponds to the low-scale system for reasoning of brain, which is fast, automatic, and intuitive type of reasoning reflects to the input information. This briefly refined information then used as the input of high--level encoder (HLE):
\begin{equation}
\quad h_{HLE} = \mathrm{GRU}_{\mathrm{HLE}}(h_{LLE}).
\end{equation}

The output feature of HLE ($h_{HLE}$) reflects the system 2 reasoning which is slower and value--guided thinking for decision making.
This LLE--to--HLE pipeline runs for $L$ reasoning cycles.

\subsection{IUE Module}
After each reasoning cycle, MHSP produces (i) a class logit vector $\ell^{(c)} \in \mathbb{R}^{B\times K}$ and (ii) an internal state $g^{(c)} \in \mathbb{R}^{B\times U}$. The IUE module assigns a scalar reliability score to each cycle. During training, it is supervised using shallow Monte--Carlo tree search (MCTS) rollouts, while at inference it provides a fast meta-cognitive estimate that guides aggregation and halting.

\subsubsection{Monte--Carlo tree search over HLE reasoning cycles}
For each cycle of HLE, we construct a search tree whose nodes correspond to cycles $c$ with states $(g^{(c)},\ell^{(c)})$ and action set $\mathcal{A}$ (e.g., ``halt" or ``continue"). During training, we run shallow MCTS simulations, and each simulation rollout yields an action flag that determines whether to halt or continue the HLE reasoning cycle. These action flags are determined with the reward value resulting from the reward head, optimized via backpropagation. We summarize the outcome at depth $c$ in a reward score $r^{(c)} \in (0,1)$ predicted by IUE module $f_{\text{IUE}}$:
\begin{equation}
r^{(c)} = \sigma\!\big(f_{\text{IUE}}([g^{(c)};\,\ell^{(c)}])\big),
\end{equation}
with targets given by the mean normalized return of simulations passing through depth $c$.

\subsubsection{MCTS--guided aggregation and adaptive halting}
At inference, we rely on the learned IUE module without running MCTS. Given per--cycle scores $\{r^{(c)}_b\}_{c=1}^{L'}$ for sample $b$, we compute attention weights:
\begin{equation}
\alpha_c(b) = \mathrm{softmax}\!\Big(\tau_{\text{ens}}\,[r^{(1)}_b,\dots,r^{(L')}_b]\Big)_c,
\end{equation}
where $\tau_{\text{ens}}$ is a temperature parameter that controls the sharpness of the attention distribution over cycles and form the final logit:
\begin{equation}
\ell_{\text{final}}(b) = \sum_{c=1}^{L'} \alpha_c(b)\,\ell^{(c)}(b).
\end{equation}

Once $c \geq 2$, if batch-mean reliability $\frac{1}{B}\sum_{b} r^{(c)}_b$ exceeds threshold $\tau_{\text{stop}}$, the model halts early.

\subsection{Training Objective}
The objective function encourages correct classification and calibrated introspection of each learning cycle. We use cross-entropy on the training logit ($\ell^{(L')}$ for the plain variant and $\ell_{\mathrm{final}}$ for IUE). A halting regularizer penalizes unnecessary cycles and excessive use of unrelated information, encouraging early stopping of HLE cycle. IUE scores $r^{(c)}$ are trained to match soft targets, reflecting correctness and confidence of interim predictions for $c{<}L'$. The total loss is the sum of cross--entropy, the halting regularizer, and IUE supervision loss, but the the halting regularizer and IUE loss are only applied when the IUE module is activated. \\

\section{EXPERIMENTS}
\subsection{Dataset}
BCI Competition IV--2a \cite{brunner2008bci} includes nine subjects (S1--S9) performing four MI tasks: left hand, right hand, both feet, and tongue. EEG signals were recorded from 22 scalp electrodes following the international 10/20 system and 3 electrooculogram (EOG) channels at 250 Hz, band--pass filtered between 0.5 and 100 Hz, and notch filtered at 50 Hz. Each subject completed two sessions on different days, each with 72 trials per class (288 trials per session). A visual cue indicated the target class, followed by the instructed imagery period. For analysis, we extracted the 3--second post--cue motor imagery interval from the EEG channels, excluding EOG channels.


\subsection{Evaluation Method}
We evaluate all models under a leave--one--subject--out (LOSO) protocol to assess subject--independent decoding. In each fold, one of the nine subjects is held out for testing and the model is trained only on the remaining eight subjects. From the training portion of each fold, 20 \% is set aside as a validation set, and we select the checkpoint with the highest validation accuracy across epochs. The best--validation model is then used for inference on the held--out subject, and performance is reported as 4--class classification accuracy. \\

\section{RESULTS AND DISCUSSION}

We evaluated three backbone architectures—EEGNet \cite{lawhern2018eegnet}, ShallowConvNet \cite{schirrmeister2017deep}, and DeepConvNet \cite{schirrmeister2017deep}—and calculate average of four--class classification accuracy and standard deviation across nine subjects (Table I). For each backbone, we compare three variants: the original baseline, the backbone augmented with the proposed MHSP module, and MHSP combined with the IUE module (MHSP+IUE). Across all backbones, the MHSP module improves average accuracy and reduces inter--subject variability. For EEGNet, accuracy increases from $0.527 \pm 0.099$ (baseline) to $0.589 \pm 0.078$ (MHSP), with MHSP+IUE yielding a further gain to $0.592 \pm 0.070$. ShallowConvNet and DeepConvNet show smaller but consistent improvements following the same trend (Table I). The IUE module also slightly tightens the across--subject standard deviation in EEGNet and DeepConvNet, suggesting that cycle--wise introspection and adaptive halting contribute to more stable performance.

Gains are particularly pronounced for difficult subjects. With EEGNet, subject S2 improves from $0.407$ to $0.532$ and S5 from $0.435$ to $0.588$ when MHSP is enabled, while high--performing subjects remain comparable. Overall, the MHSP module, and further the MHSP+IUE combination, improves average accuracy and leads to more balanced performance across subjects. \\

\section{CONCLUSIONS}
We proposed an MHSP module with an IUE module for decoding MI--based EEG signals, designed to be attached to existing backbones without modifying their core architectures. Under a LOSO protocol on the BCI Competition IV--2a dataset, the proposed components consistently improve classification performance across EEGNet, ShallowConvNet, and DeepConvNet. These gains suggest increased robustness to low signal--to--noise ratio, non--stationarity, and inter--subject variability, all of which hinder subject--independent generalization in MI--based BCI system. Future work will extend this hierarchical and meta--cognitive design to other EEG applications, such as sleep staging, single--trial event--related potential detection, and affective state decoding.

\bibliographystyle{IEEEtran}
\bibliography{MANUSCRIPT}

\end{document}